\pdfoutput=1

\documentclass[11pt]{article}

\usepackage[]{acl}

\usepackage{times}
\usepackage{latexsym}

\usepackage[T1]{fontenc}

\usepackage[utf8]{inputenc}

\usepackage{microtype}

\usepackage{booktabs}
\usepackage{indentfirst}
\usepackage{bm}
\usepackage{amsmath}
\usepackage{graphicx}
\usepackage{amssymb}
\usepackage{algorithmicx,algorithm}
\usepackage[noend]{algpseudocode}
\usepackage{multirow}
\usepackage{makecell}
\usepackage[normalem]{ulem}
\title{Overcoming Catastrophic Forgetting beyond Continual Learning: Balanced Training for Neural Machine Translation}

\author{Chenze Shao$^{1,2}$, Yang Feng$^{1,2}$\thanks{\ \ Corresponding author: Yang Feng}\\
$^{1}$ Key Laboratory of Intelligent Information Processing\\
Institute of Computing Technology, Chinese Academy of Sciences (ICT/CAS)\\
$^{2}$ University of Chinese Academy of Sciences\\
{\tt \{\href{mailto:shaochenze18z@ict.ac.cn}{shaochenze18z}, \href{mailto:fengyang@ict.ac.cn}{fengyang}\}@ict.ac.cn}}

\begin{document}
\maketitle
\begin{abstract}
Neural networks tend to gradually forget the previously learned knowledge when learning multiple tasks sequentially from dynamic data distributions. This problem is called \textit{catastrophic forgetting}, which is a fundamental challenge in the continual learning of neural networks. In this work, we observe that catastrophic forgetting not only occurs in continual learning but also affects the traditional static training. Neural networks, especially neural machine translation models, suffer from catastrophic forgetting even if they learn from a static training set. To be specific, the final model pays imbalanced attention to training samples, where recently exposed samples attract more attention than earlier samples. The underlying cause is that training samples do not get balanced training in each model update, so we name this problem \textit{imbalanced training}. To alleviate this problem, we propose Complementary Online Knowledge Distillation (COKD), which uses dynamically updated teacher models trained on specific data orders to iteratively provide complementary knowledge to the student model. Experimental results on multiple machine translation tasks show that our method successfully alleviates the problem of imbalanced training and achieves substantial improvements over strong baseline systems.\footnote{Code is available at https://github.com/ictnlp/COKD.}
\end{abstract}

\section{Introduction}
Neural Machine Translation (NMT) has achieved impressive translation performance on many benchmark datasets in the past few years \cite{cho2014learning,sutskever2014sequence,bahdanau2014neural,vaswani2017attention}. In the domain adaptation task where we have large-scale out-domain data to improve the in-domain translation performance, continual learning, which is also referred to as fine-tuning, is often employed to transfer the out-domain knowledge to in-domain \cite{luong2015stanford}. After fine-tuning, the model performs well in in-domain translation, but there is significant performance degradation in out-domain translation because it ``forgets'' the previously learned knowledge. This phenomenon is called catastrophic forgetting \cite{mccloskey1989catastrophic,french1999catastrophic} and has attracted a lot of attention \cite{goodfellow2013empirical,kirkpatrick2017overcoming,li2017learning,lee2017overcoming}.

In this work, we observe that catastrophic forgetting not only occurs in continual learning but also affects the traditional static training. To be specific, the final model pays imbalanced attention to training samples. At the end of training, the recently exposed samples attract more attention and tend to have lower losses, while earlier samples are partially forgotten by the model and have higher losses. In short, training samples receive imbalanced attention from the model, which mainly depends on the time when the model last saw the training sample (i.e., the data order of the last training epoch). 

The underlying cause of this phenomenon is mini-batch gradient descent \cite{Lecun98efficientbackprop}, that is, we do not simultaneously use all training samples to train the model but divide them into mini-batches. Therefore, training samples do not get balanced training in each update step, so we name this problem \textit{imbalanced training}. This problem is less severe in some tasks (e.g., image classification and text classification), but it has a significant impact on NMT as machine translation is a challenging task containing numerous translation rules, which are easily forgotten during the training process. Besides, we find that the imbalanced training problem is especially severe and non-negligible on low-resource machine translation.

To demonstrate that the imbalanced training problem does affect the model accuracy, we first review a widely used technique called checkpoint averaging technique, which has proved to be effective in improving model accuracy but its internal mechanisms are not fully understood. We analyze it from the perspective of catastrophic forgetting and find that their success can be attributed to the alleviation of imbalanced training. We also notice that checkpoint averaging has some limitations, leaving room for further improvements.

Inspired by the existing solution of checkpoint averaging which leverages the complementarity of checkpoints to improve model accuracy, we propose Complementary Online Knowledge Distillation (COKD) to address the problem of imbalanced training. As the model tends to forget knowledge learned from early samples, the main idea of COKD is to construct complementary teachers to re-provide this forgotten knowledge to the student. Specifically, we divide the training set into mutually exclusive subsets and reorganize them in a specific orders to train the student and teachers. We perform COKD in an online manner where teachers are on-the-fly updated to fit the need of student. When training the student on a subset, teachers can always provide the student with complementary knowledge on the other subsets, thereby preventing the student from catastrophic forgetting.

Experimental results on multiple machine translation tasks show that our method successfully alleviates the problem of imbalanced training and achieves substantial improvements over strong baseline systems. Especially, on the low-resource translation tasks that are severely affected by imbalanced training, our method is particularly effective and improves baseline models by about 2 BLEU points on average. 

In summary, our contribution is threefold:
\begin{itemize}
\item We observe the problem of \textit{imbalanced training} that training samples receive imbalanced attention from the model. We find that NMT, especially low-resource translation tasks, is seriously affected by imbalanced training.
\item We rethink the widely used checkpoint averaging technique and explain its success from the perspective of imbalanced training, which also demonstrates that the imbalanced training problem does affect the model accuracy.
\item We propose Complementary Online Knowledge Distillation for NMT, which can successfully alleviate the imbalanced training problem and improve the translation quality.
\end{itemize}
\section{Background}
\subsection{Knowledge Distillation}
Knowledge distillation \cite{hinton2015distilling} is a class of methods that transfers knowledge from a pre-trained teacher network to a student network. Assume that we are training a classifier $p(y|x;\theta)$ with $|\mathcal{V}|$ classes, and we can access the pre-trained teacher $q(y|x)$. Instead of minimizing the cross-entropy loss between the ground-truth label and the model output probability, knowledge distillation uses the teacher model prediction $q(y|x)$ as a soft target and minimizes the loss:
\begin{equation}
\label{eq:kd}
\mathcal{L}_{\text{KD}}(\theta)\!=\!-\sum_{k=1}^{|\mathcal{V}|}q(y\!=\!k|x) \times \log p(y\!=\!k|x;\theta).
\end{equation}

In neural machine translation, the standard training objective is the cross-entropy loss, which minimizes the negative log-likelihood as follows:
\begin{equation}
\label{eq:nll}
\mathcal{L}_{\text{NLL}}(\theta) = -\sum_{t=1}^{T}\log(p( y _t|y_{<t},\bm{X},\theta)),
\end{equation}
where $\bm{X}=\{x_1, ..., x_{N}\}$ and $\bm{Y}=\{y_1, ..., y_{T}\}$ are the source sentence and the target sentence, respectively. \citet{kim-rush-2016-sequence} proposed to train the student model to mimic the teacher's prediction at each decoding step, which is called Word-level Knowledge Distillation (Word-KD) and its loss is calculated as follows:
\begin{equation}
\begin{aligned}
\label{eq:wkd}
\mathcal{L}_{\text{Word-KD}}(\theta) = -&\sum_{t=1}^{T} \sum_{k=1}^{|\mathcal{V}|} q(y_{t}=k | y_{<t},\bm{X}) \times \\
&\log p(y_{t}=k | y_{<t},\bm{X},\theta).
\end{aligned}
\end{equation}

Conventional offline knowledge distillation only allows the student to learn from static pre-trained teacher models. On the contrary, online knowledge distillation trains teachers from scratch and dynamically updates them, so the student learns from different teachers during the training process. \citet{zhang2018deep} first overcame the offline limitation by training peer models simultaneously and conducted an online distillation in one-phase training between peer models. Since mutual learning requires training multiple networks, \citet{zhu2018knowledge,song2018collaborative} proposed to use a single multi-branch network for online knowledge distillation, which treats each branch as a student and the ensemble of branches as a teacher. The multi-branch architecture subsequently became the mainstream for online knowledge distillation \cite{guo2020online,chen2020online,wu2020peer}. Besides, \citet{furlanello2018born} performed iterative self-distillation where the student network is identical to the teacher in terms of the network graph. In each new iteration, under the supervision of the earlier iteration, a new identical model is trained from scratch. In NMT, \citet{wei-etal-2019-online} on-the-fly selected the best checkpoint from the training path as the teacher to guide the training process.
\begin{figure*}[t]
    \centering
    \includegraphics[width=\linewidth]{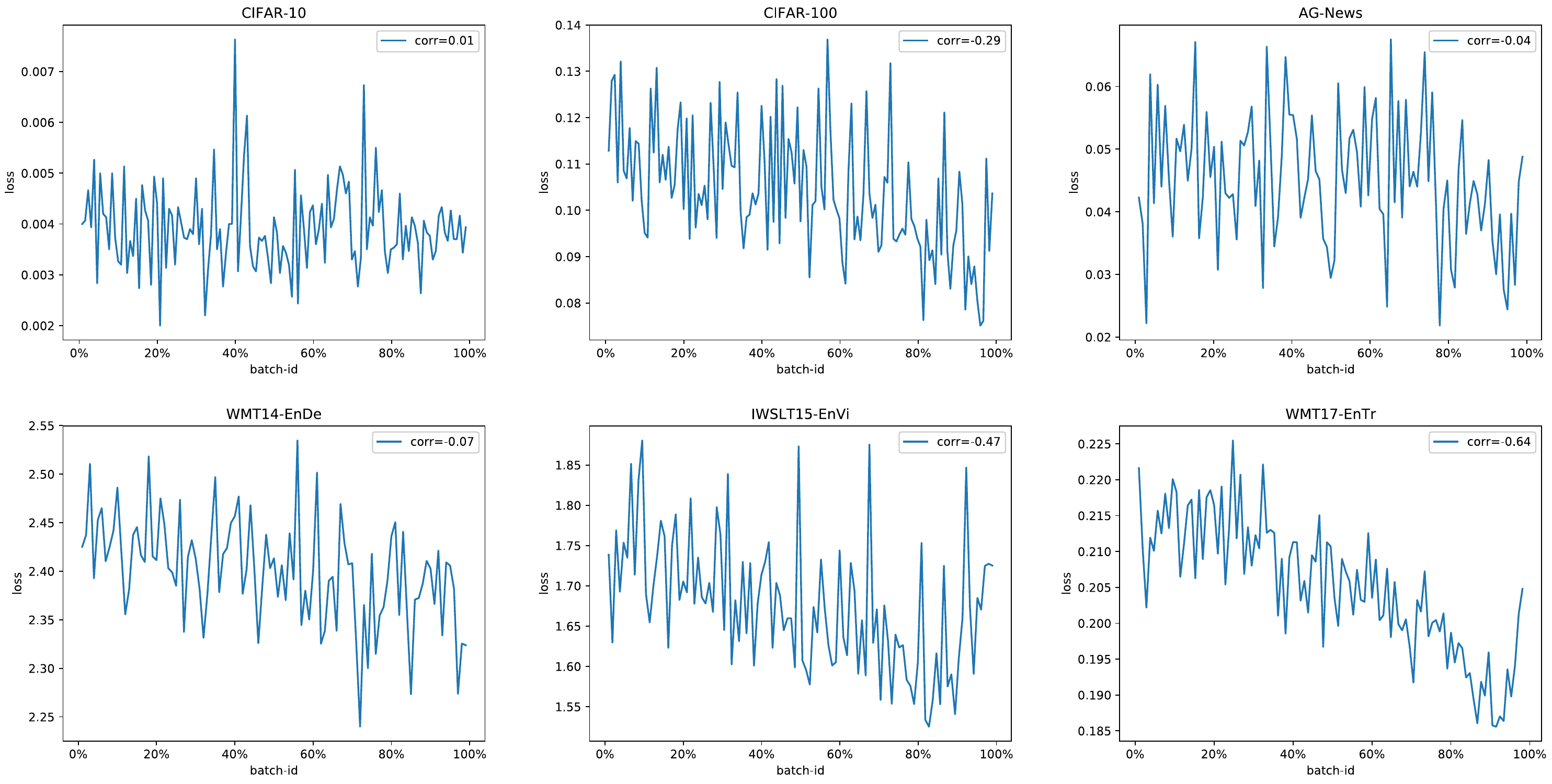}
    \caption{The relationship between the batch-id and loss on three different types of tasks. The Spearman correlation coefficient (corr) is presented in the upper right corner of charts. Batch-id $i$ indicates the i-th trained batch in the last epoch. Batch-id in the x-axis is normalized to [0,1]. Image Classification: CIFAR-10 and CIFAR-100; Text Classification: AG-News; Machine Translation: WMT14 En-De, IWSLT15 En-Vi, and WMT17 En-Tr.
 }
    \label{fig:corr}
\end{figure*}
\subsection{Catastrophic Forgetting}
Catastrophic forgetting is a problem faced by many machine learning models during continual learning, as models tend to forget previously learned knowledge when being trained on new tasks \cite{mccloskey1989catastrophic}. A typical class of methods to mitigate catastrophic forgetting is based on regularization which constrains the update of model parameters. \citet{goodfellow2013empirical} empirically find that the dropout regularization can effectively alleviate the catastrophic forgetting phenomenon. \citet{kirkpatrick2017overcoming} proposed elastic weight consolidation, which implements the modified regularization term that imposes constraints on the update of important parameters in the previous task. \citet{lee2017overcoming} proposed drop-transfer, which is a variant of dropout that drops the weight vector of turned off nodes to the weight learned on the previous task instead of a zero vector. Learning without Forgetting (LWF) \cite{li2017learning} is the approach most relevant to our work. They only use new task data to train the network but preserve the original capabilities by distilling knowledge from the pre-trained model.

There are also a number of efforts to address the catastrophic forgetting problem for the domain adaptation of NMT. \citet{kirkpatrick2017overcoming,thompson-etal-2019-overcoming} added regularization terms to constrain the update of parameters. \citet{dakwale2017finetuning} proposed to minimize the KL-divergence between the predictions of general-domain model and fine-tuned model. \citet{zeng-etal-2018-multi,gu-etal-2019-improving} introduced a discriminator to preserve the domain-shared features. \citet{liang2021finding,gu-etal-2021-pruning,xie-etal-2021-importance} fixed important parameters during the fine-tuning to preserve the general-domain performance. \citet{gu-feng-2020-investigating} investigated the cause of catastrophic forgetting from the perspectives of modules and parameters. 

\section{Imblanced Training}
\label{sec:3}
Before drawing any conclusions, we first conduct experiments on three different tasks, namely, image classification, text classification, and machine translation, to show that the problem of imbalanced training does exist. For image classification, we conduct experiments on CIFAR-10 and CIFAR-100 \cite{Krizhevsky09learningmultiple}, both of which contain 50,000/10,000 training/testing images with 32 $\times$ 32 pixels drawn from 10/100 classes. For text classification, we conduct experiments on AG-News, which contains 120,000/7,600 training/testing sentences drawn from 4 classes. For machine translation, we conduct experiments on three translation tasks: WMT14 English-German (En-De), IWSLT15 English-Vietnamese (En-Vi), and WMT17 English-Turkish (En-Tr). We use the ResNet-32 network \cite{7780459} for image classification, the VDCNN network \cite{2017Very} for text classification and Transformer-base \cite{vaswani2017attention} for machine translation. All the models are trained using cross-entropy loss. We refer readers to Appendix \ref{sec:appendix} and section \ref{sec:config} for the detailed configurations.

We train the model until convergence and then take the last checkpoint to calculate losses of training samples in the data order of the last training epoch. If there is a problem of imbalanced training, then training samples at the end of the epoch, which are recently exposed to the model, will tend to have lower losses. In contrast, training samples at the beginning will tend to have higher losses. 

For quantitative analysis, we use the Spearman correlation coefficient between the data order and loss to measure the degree of imbalanced training. Specifically, we assign each batch in the training dataset with a batch-id according to the order they appear in the last training epoch, where batch $i$ is the $i$-th trained batch. We disable regularization techniques such as dropout and label smoothing and calculate the loss for each batch. The correlation coefficient between the batch-id and the loss is used to measure the degree of imbalanced training, and a large negative correlation coefficient indicates that this problem is severe. Figure \ref{fig:corr} illustrates the relationship between the batch-id and loss. By comparing the loss curves and correlation coefficients on these six datasets, we obtain the following three main observations. 

\vspace{5pt}
\noindent{}\textbf{The problem of imbalanced training does exist.} Among the six datasets in our experiments, only CIFAR-10 has a positive correlation coefficient. Two datasets (i.e., AG-News and WMT14 En-De) have small negative correlation coefficients. Three datasets (i.e., CIFAR-100, IWSLT15 En-Vi, and WMT17 En-Tr) have an apparent decline in losses accompanied by large negative correlation coefficients. Therefore, we can conclude that the problem of imbalanced training does exist, but the degree of impact varies.

\vspace{5pt}
\noindent{}\textbf{Imbalanced training is related to task complexity.} Intuitively, imbalanced training is more likely to occur on complex tasks where previously learned knowledge may be easily forgotten during the learning of numerous new knowledge. Comparing the two image classification datasets, CIFAR-10 and CIFAR-100 have the same dataset size but a different number of classes. The correlation coefficient on the complex task CIFAR-100 is $-0.29$, while the correlation coefficient on CIFAR-10 is $0.01$. The text classification task, which only contains 4 classes, has a small correlation coefficient $-0.04$. 
Machine translation is generally considered a complex task with exponential search space and numerous translation rules. Notably, WMT17 En-Tr has the largest correlation coefficient of $-0.64$. These results are consistent with our intuition that imbalanced training has a greater impact on complex tasks like machine translation.

\vspace{5pt}
\noindent{}\textbf{Low-resource translation suffers from imbalanced training.} Comparing the three machine translation datasets, the imbalanced training problem has a much larger impact on low-resource datasets (i.e., IWSLT15 En-Vi and WMT17 En-Tr), where the high-resource dataset WMT14 En-De is less affected. To eliminate the influence of language, we randomly select 100K sentences from the WMT14 En-De dataset for the training to simulate the low-resource scenario. We show the loss curve in Appendix \ref{sec:applowende}, where the corresponding correlation coefficient is $-0.63$, which also supports the conclusion. This is counter-intuitive since when there are many training samples, the early samples seem to be more easily forgotten. Actually, as Figure \ref{fig:corr} shows, the loss curves are generally less steep at the beginning, indicating that early samples are nearly ``equally forgotten'' by the model. For high-resource datasets, most samples are nearly ``equally forgotten'' and only the losses of a few samples at the end are highly correlated with the batch-id, so the overall correlation is low.  In comparison, nearly the whole loss curve of low-resource datasets is steep, so the model may simultaneously overfit recent samples and underfit early samples due to imbalanced training. Therefore, the problem of imbalanced training is more serious and nonnegligible in low-resource machine translation. 

\vspace{5pt}
\noindent{}\textbf{Loss rises in the end due to the momentum of optimizer.} On CIFAR-100, IWSLT15 En-Vi, and WMT17 En-Tr, though their loss curves are generally downward, they all have a sudden rise at the end. This abnormal phenomenon is actually consistent with our conclusion. Because of the momentum factor in the adam optimizer, the impact of a model update is not limited to the current step. The optimizer retains the gradient in the form of momentum, which will affect the gradient updates in the next few steps. Therefore, the impact of momentum is not fully released in the last few training steps, so the loss rises in the end. 

\section{Checkpoint Averaging}
Checkpoint averaging, which directly takes the average of parameters of the last few checkpoints as the final model, is a widely used technique in NMT \cite{junczys-dowmunt-etal-2016-amu,vaswani2017attention}. The averaged checkpoint generally performs better than any single checkpoint. However, to the best of our knowledge, its internal mechanism is not fully understood.

In this section, we analyze the success of checkpoint averaging from the perspective of imbalanced training. Though training samples receive imbalanced attention from each checkpoint, this imbalance is different among checkpoints. If we understand the imbalanced training as the noise on each checkpoint, noises among different checkpoints can be approximately regraded as i.i.d. random variables. By averaging checkpoints, the variance of random noise is reduced and thereby alleviating the problem of imbalanced training. Based on the above analysis, we make the following hypothesis and verify it through experiments.
\paragraph{Hypothesis} \emph{Checkpoint averaging improves the model performance through alleviating the problem of imbalanced training.}
\paragraph{Experiments} We conduct experiments on the six datasets to study the relationship between checkpoint averaging and imbalanced training. We average the last five epoch checkpoints and compare their performance with the best single checkpoint. Table \ref{tab:ave} reports the model performance along with the correlation coefficient on the six datasets. We can see that checkpoint averaging achieves considerable improvements on datasets where the problem of imbalanced training is severe. On datasets with small correlation coefficients, the improvements of checkpoint averaging are very limited. These results confirm our hypothesis and also demonstrate that the imbalanced training problem does affect the model accuracy.

\begin{table}[ht]
\centering
\begin{tabular}{c|c|c|c}
\toprule
{\bf Dataset }& {\bf Corr}&{\bf Best}&{\bf Ave}\\
\hline \multicolumn{4}{c}{\em Image Classification \& Text Classification} \\ \hline
CIFAR-10&0.01&93.51\%&93.47\%\\
CIFAR-100&-0.29&70.89\%&71.36\%\\
AG-News&-0.04&91.61\%&91.70\%\\
\hline \multicolumn{4}{c}{\em Machine Translation} \\ \hline
WMT14 En-De&-0.07&27.29&27.45\\
IWSLT15 En-Vi&-0.47&28.52&29.08\\
WMT17 En-Tr&-0.64&12.79&13.42\\
\bottomrule
\end{tabular}
\caption{Model performance on the test sets of six datasets. For classification tasks, we report the Top-1 accuracy. For translation tasks, we report the BLEU score. \textbf{Corr} is the Spearman correlation coefficient calculated in section \ref{sec:3}. \textbf{Best} and \textbf{Ave} represent the best and average checkpoint performance, respectively.}
\label{tab:ave}
\end{table}
\paragraph{Limitations} Though checkpoint averaging can alleviate the problem of imbalanced training and improve the model performance, it also has some limitations and its success largely depends on the empirical choice of checkpoint interval. If the checkpoint interval is small, then the i.i.d. assumption does not hold, so the imbalance cannot be effectively eliminated and may even become stronger (Appendix \ref{sec:appave}). If the checkpoint interval is large, then checkpoints may not lie in the same parameter space, making the direct averaging of checkpoints problematic.

\section{Approach}
In this section, we propose Complementary Online Knowledge Distillation (COKD) to alleviate the problem of imbalanced training. We apply knowledge distillation with dynamically updated complementary teachers to re-provide the forgotten knowledge to the student model.

\subsection{Complementary Teachers}
We first introduce the construction of complementary teachers. Assume that we have $n$ teacher models $\mathcal{T}_{1:n}$ and the student model is $\mathcal{S}$, and both teacher models and the student model are randomly initialized. We expect that teacher models should be dynamically updated so that they are always complementary to the student. While the student learns from new training samples and gradually forgets early samples, teacher models should re-provide the forgotten knowledge to the student.

Recall that the model pays imbalanced attention to different training samples depending on the data order of the training. Therefore, a natural way to obtain complementary teachers is to train teachers in different data orders. Specifically, in each epoch, we divide the training dataset $\mathcal{D}$ into $n\!+\!1$ mutually exclusive splits $(\mathcal{D}_{1},\mathcal{D}_{2},...,\mathcal{D}_{n+1})$. The student model sequentially learns from $\mathcal{D}_{1}$ to $\mathcal{D}_{n+1}$, where the data order is different for teacher models.

We use a ordering function $\mathcal{O}(i,t)$ to denote the training data for teacher $\mathcal{T}_{i}$ at time $t$. After teacher models $\mathcal{T}_{1:n}$ learn from data splits $\mathcal{D}_{\mathcal{O}(1:n,t)}$ respectively, the student $\mathcal{S}$ learn from both $\mathcal{D}_{t}$ and teachers.
To make teachers complementary with the student, the ordering function $\mathcal{O}(\cdot,t)$ should cover all data splits except $\mathcal{D}_{t}$. To ensure that each teacher can access the whole training data, the ordering function $\mathcal{O}(i,\cdot)$ should also cover all data splits. Fortunately, we find that a simple assignment of $\mathcal{O}$ satisfies the above requirements:
\begin{equation}
\mathcal{O}(i,t)= \left\{
             \begin{array}{ll}
             i+t,&i+t \leq n+1 \\
             i+t\!-\!n\!-\!1,&i+t > n+1
             \end{array}
             .\right.
\end{equation}
where $i\in \{1,2,...,n\}$ and $t\in \{1,2,...,n+1\}$. Under this assignment, teacher $\mathcal{T}_{i}$ simply uses the data split that has offset $i$ from the student, which ensures that all teachers are complementary with the student and can access the whole training set.

\subsection{Complementary Training}
The knowledge of $n$ complementary teachers can be transfered to the student through word-level knowledge distillation:
\begin{equation}
\begin{aligned}
\label{eq:word-kdm}
\mathcal{L}_{\text{KD}}(\theta)\! =\! &-\!\sum_{t=1}^{T} \sum_{k=1}^{|\mathcal{V}|}\sum_{i=1}^{n}\frac{q_i(y_{t}=k | y_{<t},\bm{X})}{n} \\
&\times  \log p(y_{t}=k | y_{<t},\bm{X},\theta),
\end{aligned}
\end{equation}
where $p$ is the prediction of student $\mathcal{S}$ and $q_i$ is the prediction of teacher $\mathcal{T}_{i}$. We use a hyperparameter $\alpha$ to interpolate the distillation loss and the cross-entropy loss:
\begin{equation}
\label{eq:word-inter}
\mathcal{L}(\theta)=\alpha \cdot \mathcal{L}_{\text{KD}}(\theta)+ (1-\alpha)\cdot\mathcal{L}_{\text{NLL}}(\theta).
\end{equation}

In this way, the student model learns both new knowledge from the training set and complementary knowledge from teacher models. With an appropriate $\alpha$, we can achieve a balance between the two kinds of knowledge and alleviate the problem of imbalanced training. However, this method is based on knowledge distillation where knowledge is transferred unidirectionally from teachers to the student. Though the student can benefit from balanced training, these complementary teachers also set an upperbound to the student and prevent it from performing better. 

To overcome this limitation, we follow the underlying idea of two-way knowledge transfer where the knowledge is also transferred from the student to teachers \cite{zhang2018deep,zhu2018knowledge}. We use a simple reinitialization method to achieve the two-way knowledge transfer. At the end of each epoch, we reinitialize teacher models with the parameters of the student model:
\begin{equation}
\label{eq:init}
\mathcal{T}_{i} \leftarrow \mathcal{S},\ \ \ i \in \{1,2,...,n\}.
\end{equation}
Through the reinitialization, the student and teachers are exactly the same at the beginning of each epoch. In this way, both the student and teachers are iteratively improved so the student performance is no longer limited by the fixed ability of teachers. We summarize the training process of COKD in Algorithm \ref{alg}.

\begin{algorithm}[t]
\caption{COKD} 
\label{alg}
\hspace*{0.02in} {\bf Input:} training set $\mathcal{D}$, the number of teachers $n$\\
\hspace*{0.02in} {\bf Output:} 
student model $\mathcal{S}$
\begin{algorithmic}[1]
\State randomly initialize student $\mathcal{S}$ and teachers $\mathcal{T}_{1:n}$
\While{not converge}
\State randomly divide $\mathcal{D}$ into $n\!+\!1$ subsets $(\mathcal{D}_{1},\mathcal{D}_{2},...,\mathcal{D}_{n+1})$
\For{$t = 1$ {\bf{to}} $n+1$}
\For{$i = 1$ {\bf{to}} $n$}
\State train $\mathcal{T}_{i}$ on $\mathcal{D}_{\mathcal{O}(i,t)}$
\EndFor
\State train $\mathcal{S}$ on $\mathcal{D}_{t}$ according to Equation \ref{eq:word-inter}
\EndFor
\For{$i = 1$ {\bf{to}} $n$}
$\mathcal{T}_{i}$ $\leftarrow$ $\mathcal{S}$
\EndFor
\EndWhile
\State \Return student model $\mathcal{S}$
\end{algorithmic}
\end{algorithm}

\section{Experiments}
\subsection{Setup}
\label{sec:config}
To evaluate the performance of COKD, we conduct experiments on multiple machine translation tasks. For low-resource translation where the problem of imbalanced training is severe, we run experiments on WMT17 English-Turkish (En-Tr, 207K sentence pairs), IWSLT15 English-Vietnamese (En-Vi, 133K sentence pairs), and TED bilingual dataset. We also evaluate the high-resource performance of COKD on WMT14 English-German (En-De, 4.5M sentence pairs). For WMT17 En-Tr and IWSLT15 En-Vi, we use case-sensitive SacreBLEU \cite{post-2018-call} to report reproducible BLEU scores. For TED bilingual dataset, following \citet{xu-etal-2021-vocabulary}, we report the tokenized BLEU. For WMT14 En-De translation, we report the tokenized BLEU \cite{papineni2002bleu} with compound split.

For WMT17 En-Tr, we use \textit{newstest2016} as the validation set and \textit{newstest2017} as the test set. We learn a joint BPE model \cite{sennrich-etal-2016-neural} with 16K operations. For IWSLT15 En-Vi, we use the pre-processed data used in \citet{luong2015stanford}\footnote{https://github.com/tefan-it/nmt-en-vi}. For TED bilingual dataset, we use the pre-processed data used in \citet{xu-etal-2021-vocabulary}\footnote{https://github.com/Jingjing-NLP/VOLT}. For WMT14 En-De, the validation set is \textit{newstest2013} and the test set is \textit{newstest2014}. We learn a joint BPE model with 32K operations.

In the main experiments, we set the number of teachers $n$ to 1 and the hyperparameter $\alpha$ to 0.95. We implemented our approach based on the base version of Transformer \cite{vaswani2017attention}. Following \citet{wei-etal-2019-online}, we increase the dropout rate to 0.2 on WMT17 En-Tr and IWSLT15 En-Vi. For TED bilingual dataset, we further increase the dropout rate of Transformer baseline to 0.3. All models are optimized with Adam \cite{DBLP:journals/corr/KingmaB14} with the optimizer settings in \citet{vaswani2017attention}. The batch size is 32K for all translation tasks. For inference, we average the last 5 checkpoints and use beam search with beam size 5. The checkpoint interval is 1000 for low-resource tasks and 5000 for WMT14 En-De.

\subsection{Main Results}
We first conduct experiments on the two low-resource datasets WMT17 En-Tr and IWSLT15 En-Vi and the high-resource dataset WMT14 En-De to evaluate the capability of our method. We compare our method with knowledge distillation methods and deep mutual learning \cite{zhang2018deep}, and also report the results of Online Distillation from Checkpoints (ODC) \cite{wei-etal-2019-online} for comparison. The results are listed in Table \ref{tab:main}.

\vspace{5pt}
\noindent\textbf{Low-Resource} First, we focus on the results on the two low-resource datasets where the problem of imbalanced training is severe. Since we have applied the checkpoint averaging technique on the baseline system, our baseline is very competitive and outperforms the baseline of \citet{wei-etal-2019-online}. We refer readers to Appendix \ref{app:noave} for results without checkpoint averaging. Knowledge distillation techniques and deep mutual learning bring some improvements to the baseline, but the improvements are relatively weak. In comparison, COKD substantially improves the baseline performance by about 3 BLEU scores, demonstrating the effectiveness of COKD on low-resource translation tasks. 

{
\centering
\begin{table}[t]
  \small
  \begin{center}
    \begin{tabular}{c|ccc}
    \toprule
     \bf Models &\bf{ En-Tr}&\bf{ En-Vi}&\bf{En-De} \\
    \midrule
    Transformer$^*$ & 12.20 & 28.56 & --\\
    ODC$^*$  & 12.92 & 29.47 & --\\
  \midrule
    Transformer   &13.42& 29.08&27.45\\
    Word-KD  &13.66 &29.54&27.76\\
    Seq-KD  &13.91&29.69&27.84\\
    Mutual&13.72&29.83&27.81\\
 \midrule
    COKD&\textbf{16.66}&\textbf{31.95}&\textbf{28.26}\\
    \bottomrule
  \end{tabular}
  \caption{BLEU scores on three translation tasks. $^*$ means results reported in \citet{wei-etal-2019-online}. Word-KD means word-level knowledge distillation, and Seq-KD means sequence-level knowledge distillation. Mutual means our reimplementation of deep mutual learning \cite{zhang2018deep}.}
  \label{tab:main}
  \end{center}
\end{table}
}

\vspace{5pt}
\noindent\textbf{High-Resource}
On the high-resource dataset WMT14 En-De, COKD still outperforms the baseline and knowledge distillation methods. The improvement of COKD is relatively small compared to the low-resource setting, which can be explained from the perspective of imbalanced training. As illustrated in Figure \ref{fig:corr}, high-resource datasets like WMT14 En-De is less affected by the problem of imbalanced training, so the alleviation of this problem may not bring strong improvements on high-resource datasets.

\vspace{5pt}
\noindent\textbf{TED Bilingual Dataset} We further conduct experiments on TED bilingual dataset to confirm the effectiveness of COKD on low-resource translation tasks. We evaluate COKD on both En-X and X-En directions and report the results in Table \ref{tab:ted}. The performance of COKD is still very impressive, which improves the baseline by 1.59 BLEU on average in the En-X direction, and improves the baseline by 2.15 BLEU on average in the En-X direction.

\begin{table*}[t]
    \centering
    \footnotesize
    \begin{tabular}{c|ccccccccccccc}
    \toprule
    \multirow{1}{*}{En-X}    &  Es &  PTbr &  Fr &  Ru & He &  Ar &    It &  Nl &  Ro &  Tr  &  De & Vi & Ave \\ \midrule 
       
        Base  & 40.86 & 40.31 & 41.27 & 21.86  & 29.01 &18.40 & 36.37 & 34.06 & 28.33 & 17.70 & 31.46 & 29.66& 30.77\\
        COKD  &   42.50 &   42.46 &   43.15 &    22.94 &   30.22 &    19.36    &   37.78  &    35.87  &     29.70  &    19.50 &   33.48 &    31.33 &32.36\\
    \toprule
      X-En   &  Es &  PTbr & Fr & Ru &  He &  Ar  & It &  Nl &  Ro &  Tr  &  De & Vi & Ave\\ \midrule 
      
        Base &   42.94 & 45.52 & 41.32 & 26.21 & 38.78 & 33.06 & 39.55 & 37.52 & 36.50 & 27.19 & 36.89 & 27.64 &36.09\\
        COKD     &  44.72 &   47.84 &   43.33  &   27.87 &    40.81 &   35.03 &   41.48 &   39.66 &   38.78 &   29.68 &   39.73 &   29.91 &38.24\\
       \bottomrule
    \end{tabular}
        \caption{BLEU scores on the TED bilingual dataset. Ave means the average BLEU.}
    \label{tab:ted}
\end{table*}
\subsection{Ablation Study}
In this section, we study the effect of complementary teachers and teacher reinitialization in COKD. We remove each of them respectively and report their performance in Table \ref{tab:abla}. By removing complementary teachers, we do not split the dataset and assign random data order to teachers, which leads to obvious performance degradtion. We also notice that a large part of improvement comes from the reinitialization, suggesting the importance of two-way knowledge transfer where both the student and teachers are iteratively improved.
\begin{table}[t]
  \small
  \begin{center}
    \begin{tabular}{l|ccc}
    \toprule
     \bf Models &\bf{ En-Tr}&\bf{ En-Vi}&\bf{En-De} \\
    \midrule
    COKD&16.66&{31.95}&{28.26}\\
    \ \ - CT&15.83 &31.56&27.96\\
    \ \ - TR&14.02&29.93&27.84\\
    \bottomrule
  \end{tabular}
  \caption{Ablation study for COKD. CT means complementary teachers. TR means teacher reinitialization.}
  \label{tab:abla}
  \end{center}
\end{table}

\subsection{Hyperparameters}
There are two hyperparameters in COKD: the number of teachers $n$ and the loss weight $\alpha$, whose default settings are $n=1$, $\alpha=0.95$ in the main experiments. In this section, we conduct experiments on WMT17 En-Tr to show the effect of the two hyperparameters.

\vspace{5pt}
\noindent{}\textbf{The number of teachers} We change the number of teachers $n$ from 1 to 5 to evaluate the effect of $n$ in COKD and report the BLEU score and training time in Table \ref{table:n}. We find that using more teachers does not necessarily lead to better performance, suggesting that the main improvement is not due to the ensemble of multiple teachers. Large $n$ may slightly outperform the $n=1$ setting but comes with a larger training cost. Therefore, we recommend the $n=1$ setting in practical applications. Though the training cost is still larger, it is acceptable on low-resource datasets considering the strong performance improvement.

\begin{table}[t]
\small
\centering
\begin{tabular}{c|ccccc}
\toprule
$n$&0&1&2&3&4\\
\midrule
BLEU&12.57&15.43&15.35&15.66&15.47\\
\midrule
Time&1.34h&2.87h&4.56h&6.15h&7.83h\\
\bottomrule
\end{tabular}
\caption{BLEU scores of COKD on the validation set of WMT17 En-Tr. The training time is measured with 8 NVIDIA RTX 2080Ti. h is the abbreviation of hour.}
\label{table:n}
\end{table}
\vspace{5pt}
\noindent{}\textbf{Hyperparameter $\alpha$} We set the hyperparameter $\alpha$ to $0.75, 0.9, 0.95, 0.98, 1$ respectively. The corresponding BLEU scores are listed in Table \ref{table:alpha}. We can see that the model performance is sensitive to the hyperparameter $\alpha$. Generally, the model prefers large $\alpha$, where a slightly smaller $\alpha$ may significantly degrade the model performance. We explain this phenomenon as the imbalance of complementary knowledge and new knowledge. The distillation loss carries the complementary knowledge and the cross-entropy loss carries the new knowledge, so an appropriate $\alpha$ should balance the two kinds of knowledge. Considering that the distillation loss is only a little biased to the complementary knowledge, $\alpha$ should be much larger than $0.5$, otherwise it cannot keep the balance. We empirically recommend the $\alpha=0.95$ setting, which also shows good performance on other datasets.

\begin{table}[t]
\small
\centering
\begin{tabular}{c|c|c|c|c|c}
\toprule
$\alpha$&0.75&0.9&0.95&0.98&1\\
\midrule
BLEU&13.86&14.94&15.43&15.31&15.23\\
\bottomrule
\end{tabular}
\caption{BLEU scores of COKD with different $\alpha$ on the validation set of WMT17 En-Tr.}
\label{table:alpha}
\end{table}

\subsection{COKD Alleviates Imbalanced Training}
\begin{figure}[t]
    \centering
    \includegraphics[width=1\linewidth]{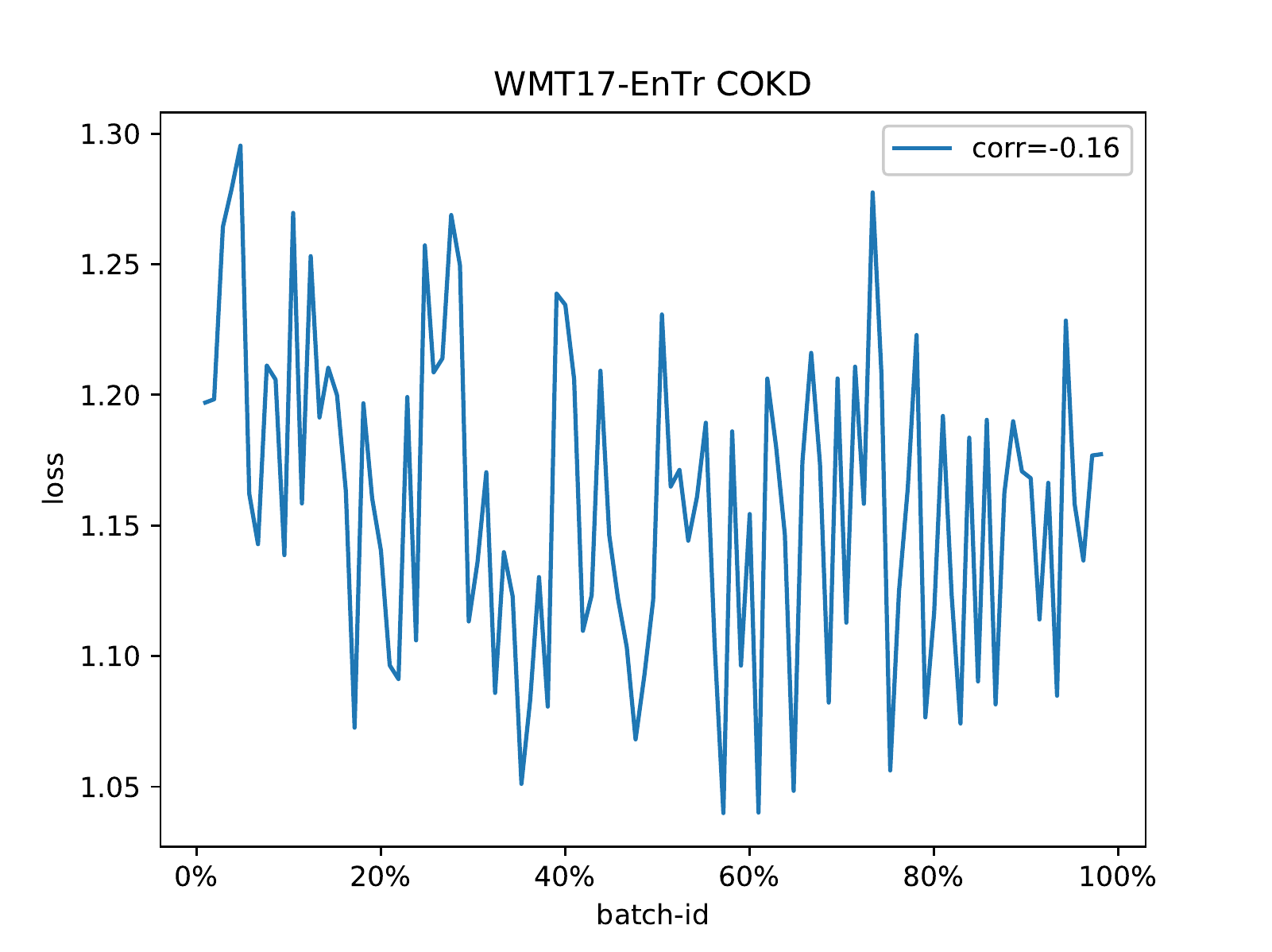}  
    \caption{The loss curve of COKD on the last epoch.}
    \label{fig:cokd}
\end{figure}

In this section, we evaluate the effectiveness of COKD in alleviating the problem of imbalanced training. We take the final model of COKD and measure the correlation between batch-id and loss in the last epoch. We conduct experiments on the WMT17 En-Tr dataset where the problem of imbalanced training is severe. As Figure \ref{fig:cokd} shows, the downward trend of loss is successfully alleviated by COKD, and the correlation coefficient is improved from $-0.64$ to $-0.16$.

\section{Conclusion}
In this paper, we observe that catastrophic forgetting will cause imbalanced training, which is severe in low-resource machine translation and will affect the translation quality. We rethink the checkpoint averaging technique and explain its success from the perspective of imbalanced training. We further propose Complementary Online Knowledge Distillation (COKD), which successfully alleviates the imbalanced training problem and achieves substantial improvements in translation quality. 
\section{Acknowledgement}
We thank the anonymous reviewers for their insightful comments. This work was supported by National Key R\&D Program of China (NO.2017YFE9132900).
\bibliography{acl_latex}
\bibliographystyle{acl_natbib}

\appendix
\onecolumn
\section{Configurations for Classification Tasks}
\label{sec:appendix}
In this section, we describe the configurations of classification tasks in detail. For the two image classification tasks CIFAR-10 and CIFAR-100, our implementation is based on the source code released by \citet{chen2020online}\footnote{https://github.com/DefangChen/OKDDip-AAAI2020}. We use the ResNet-32 network \cite{7780459}. For preprocessing, we normalized all images by channel means and standard deviations. We use stochastic gradient descent with Nesterov momentum for optimization and set the initial learning rate to 0.1, momentum to 0.9. We set the mini-batch size to 128 and weight decay to 5$\times$10-4. The learning rate is divided by 10 at 150 and 225 of the total 300 training epochs for these two datasets.

For the text classification task AG-News, our implementation is based on an open-source NLP benchmark\footnote{https://github.com/ArdalanM/nlp-benchmarks}. We use the VDCNN network \cite{2017Very} with depth 29. We use stochastic gradient descent with momentum for optimization and set the initial learning rate to 0.01, momentum to 0.9. We train the model for 100 epochs and multiply the learning rate by 0.9 every 15 steps. We set the mini-batch size to 128. 
\section{Simulation of Low-Resource WMT14 En-De}
\begin{figure}[h]
    \centering
    \includegraphics[width=0.6\linewidth]{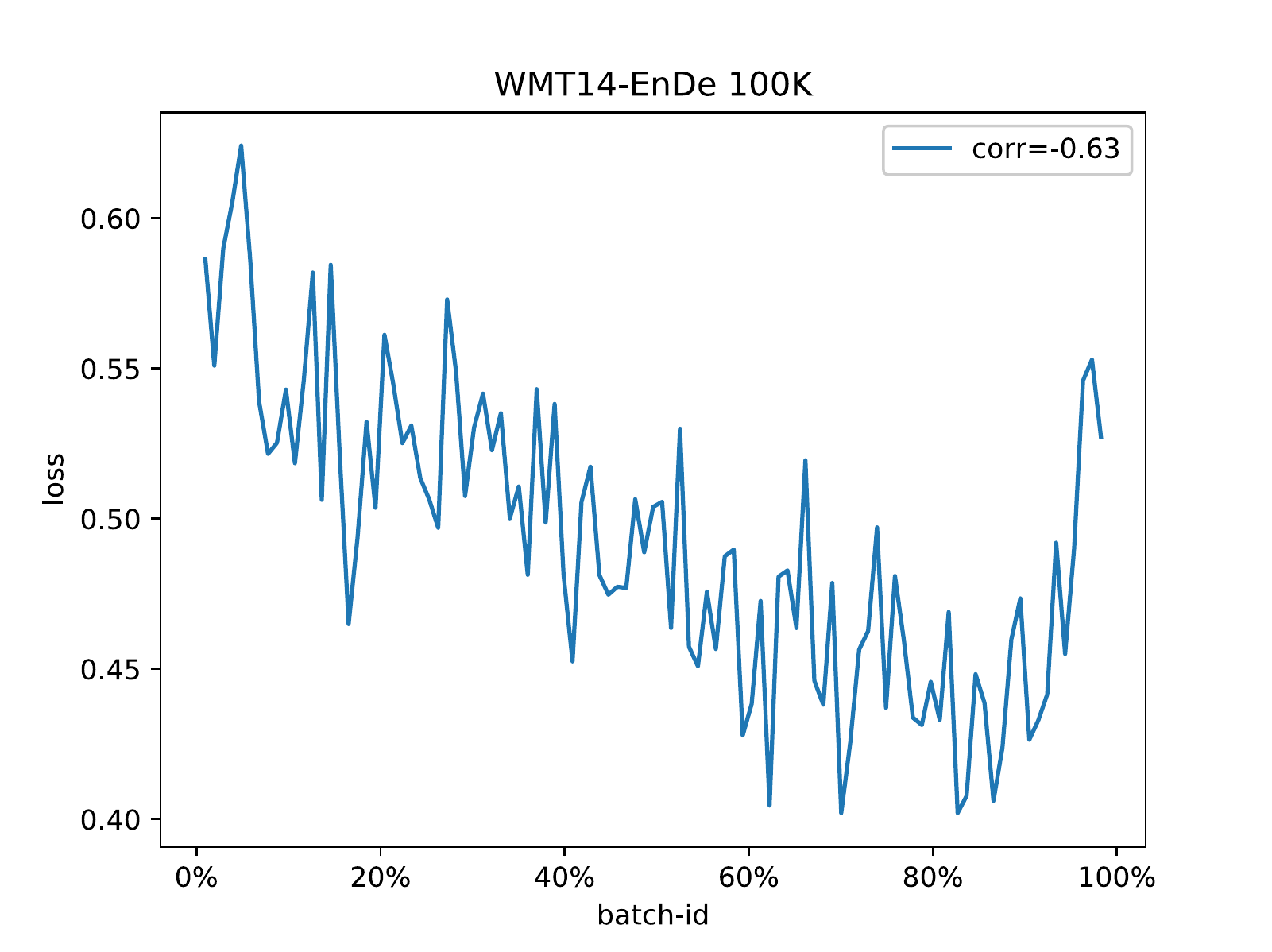}
    \caption{Relationship between the batch-id and loss on the simulated low-resource scenario of WMT14 En-De.}
    \label{fig:sim}
\end{figure}

We randomly select 100K sentences from the WMT14 En-De dataset for the training to simulate the low-resource scenario. Figure \ref{fig:sim} shows that the loss curve is downward and has a large negative correlation coefficient $-0.63$. Comparing with the whole dataset result, it confirms the conclusion that low-resource translation suffers more from imbalanced training.
\label{sec:applowende}
\label{sec:appother}

\section{Case Analysis of Checkpoint Averaging}
\label{sec:appave}
\begin{figure}[h]
    \centering
    \includegraphics[width=0.6\linewidth]{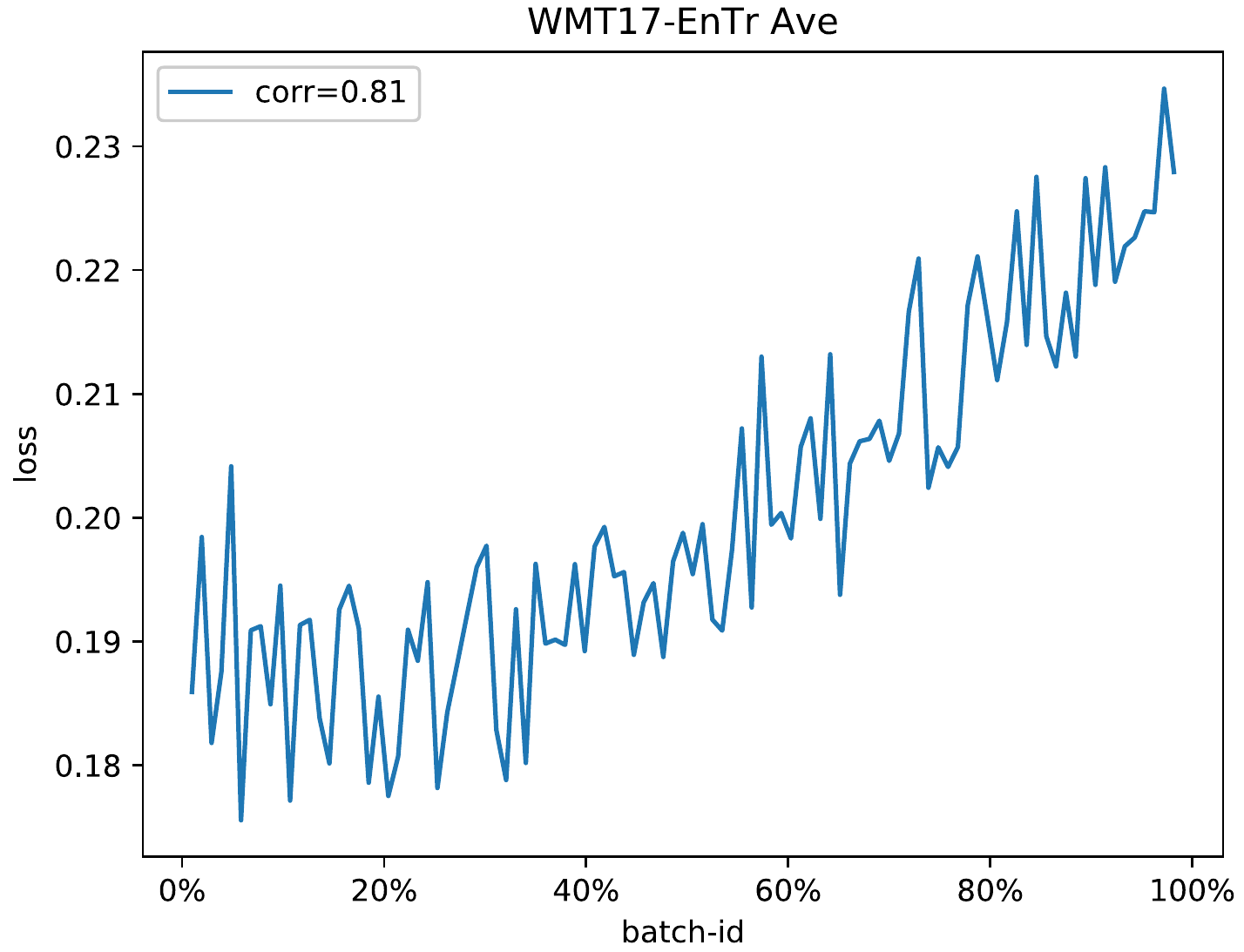}
    \caption{Relationship between the batch-id and loss on WMT17 En-Tr.}
    \label{fig:ave}
\end{figure}

In this section, we give a case analysis to show that checkpoint averaging may bring new imbalance and degrade the model performance. We conduct experiments on the validation set of WMT17 En-Tr translation. We set the checkpoint interval to $0.1$ epoch, so $10$ checkpoints are saved in the last training epoch. We average these $10$ checkpoints and illustrate the relationship between the batch-id and loss in Figure \ref{fig:ave}.

Different from the previous imbalance, Figure \ref{fig:ave} shows an upward trend with a large positive correlation coefficient of 0.81. The BLEU score of the averaged model is 12.71, which is even lower than the BLEU of a single checkpoint. Intuitively, this is because earlier samples are recently exposed to many checkpoints, while the latter samples are only exposed to the last few checkpoints. The underlying cause is the small interval of checkpoints. As the checkpoint interval is small, the i.i.d. assumption does not hold, so checkpoint averaging cannot eliminate the imbalance and may bring a new imbalance.
\section{Performance without Checkpoint Averaging}
\label{app:noave}
In the main experiments, we still apply the checkpoint averaging technique during inference to alleviate the imbalanced training problem. Here we report the performance without checkpoint averaging in Table \ref{tab:noave}.

{
\centering
\begin{table}[h]
  \small
  \begin{center}
    \begin{tabular}[b]{cccccccccc}
    \toprule
     \multirow{2}{*}{\textbf{Models}} &
 \multicolumn{3}{c}{\textbf{En-Tr}} & \multicolumn{3}{c}{\textbf{En-Vi}}& \multicolumn{3}{c}{\textbf{En-De}} \\
 &{ Best}&{ Ave}&{$\Delta$}&{ Best}&{ Ave}&{$\Delta$}&{ Best}&{ Ave}&{$\Delta$}\\
\midrule
    Transformer &12.79&13.42&+0.63&28.52&29.08&+0.56&27.29&27.45&+0.16\\
    Word-KD &13.25&13.66&+0.41&29.09&29.54&+0.45&27.64&27.76&+0.12\\
    Seq-KD &13.37&13.91&+0.54&29.26&29.69&+0.43&27.65&27.84&+0.19\\
    Mutual &13.13 &13.72 & +0.59&29.35&29.83&+0.48&27.59&27.81&+0.22\\
    COKD &16.61&16.66&+0.05&31.86&31.95&+0.09&28.18&28.26&+0.08\\
    \bottomrule
  \end{tabular}
  \caption{BLEU scores on three translation dateset. \textbf{Best} and \textbf{Ave} represent the best and average checkpoint performance, respectively.}
  \label{tab:noave}
  \end{center}
\end{table}
}

We can see that the performance of COKD only decreases a little after removing checkpoint averaging, indicating that COKD itself can successfully alleviate the problem of imbalanced training. In contrast, the performance gap between Best and Ave is larger on other NMT Systems, suggesting that the problem of imbalanced training is severe and cannot be simply mitigated by other techniques like knowledge distillation and mutual learning.
\section{Fine-Tuning with Lower Learning Rate}
When the training is finished, we can manually reduce the learning rate to fine-tune the model, which also improves the model accuracy in some situations. The common explanation is that fine-tuning can help to converge the optimization process and reduce the loss function. Here we analyze fine-tuning from another perspective and find that it can also alleviate the imbalanced training problem.

\begin{figure}[h]
    \centering
    \includegraphics[width=0.7\linewidth]{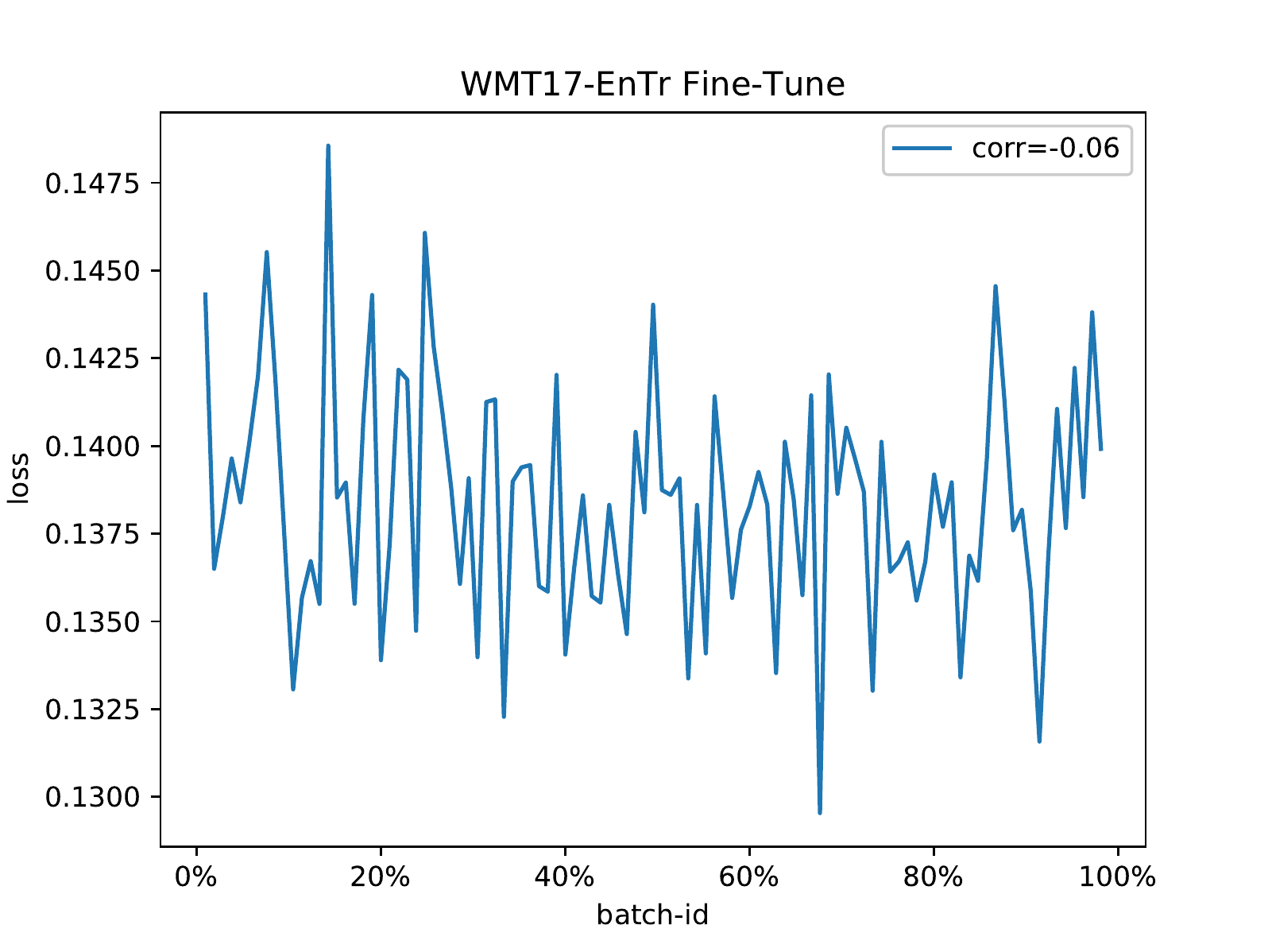}
    \caption{Relationship between the batch-id and loss on WMT17 En-Tr.}
    \label{fig:lr}
\end{figure}

Intuitively, if the learning rate is very low, the model will not be much biased towards recent training samples, which should reduce the imbalance. We conduct experiments on WMT17 En-Tr to confirm our hypothesis. We reduce the learning rate from the base setting of $7\cdot10^{-4}$ to $1\cdot10^{-5}$ to fine-tune the model and draw the loss curve in Figure \ref{fig:lr}. From both the correlation coefficient and loss curve, we can see that the imbalance is greatly reduced by fine-tuning. Regrading the BLEU score, the BLEU score after fine-tuning is only 0.25 higher than the baseline, which is much smaller than the improvement of checkpoint averaging. We speculate that it is due to some drawbacks of fine-tuning. For example, fine-tuning with a low learning rate has a risk of overfitting the dataset, which may influence its performance on low-resource datasets.

\end{document}